\documentclass[final]{cvpr}

\usepackage[utf8]{inputenc}
\usepackage{times}
\usepackage{epsfig}
\usepackage{graphicx}
\usepackage{amsmath}
\usepackage{breqn}
\usepackage{amssymb}
\usepackage{multirow}
\usepackage{booktabs}
\usepackage{hhline}
\usepackage{subcaption}
\usepackage{amsmath, bm}

\usepackage[pagebackref=true,breaklinks=true,colorlinks,bookmarks=false]{hyperref}

\newcommand{\Modelname}{SMPLicit}

\renewcommand{\vec}[1]{\boldsymbol{#1}}
\newcommand{\mat}[1]{\mathbf{#1}}

\newcommand{\pose}[0]{\vec{\theta}}
\newcommand{\shape}[0]{\vec{\beta}}

\newcommand{\cut}{\mat{z}_\mathrm{cut}}
\newcommand{\style}{\mat{z}_\mathrm{style}}
\newcommand{\posenc}{\mat{P}_{\shape}}
\newcommand{\zpose}{\mat{z}_{\pose}}

\def\cvprPaperID{4685} 
\def\confYear{CVPR 2021}

\begin{document}

\title{\Modelname : Topology-aware Generative Model for Clothed People}

\author{\hspace{-5mm}Enric Corona$^{1}$ \hspace{1.5mm}
Albert Pumarola$^{1}$ \hspace{1.5mm} Guillem Aleny\`a$^{1}$ \hspace{1.5mm} Gerard Pons-Moll$^{2,3}$ \hspace{1.5mm} Francesc Moreno-Noguer$^{1}$\\
\hspace{-4mm}${}^{1}$Institut de Robòtica i Informàtica Industrial, CSIC-UPC, Barcelona, Spain\\${}^{2}$University of T\"{u}bingen, Germany, ${}^{3}$Max Planck Institute for Informatics, Germany}

\twocolumn[{%
\renewcommand\twocolumn[1][]{#1} %
\maketitle
\thispagestyle{empty}
\begin{center}
    \centering
    \vspace{-6mm}
    \includegraphics[width=.95\linewidth, trim={0cm 0.3cm 0cm 0}, clip = true]{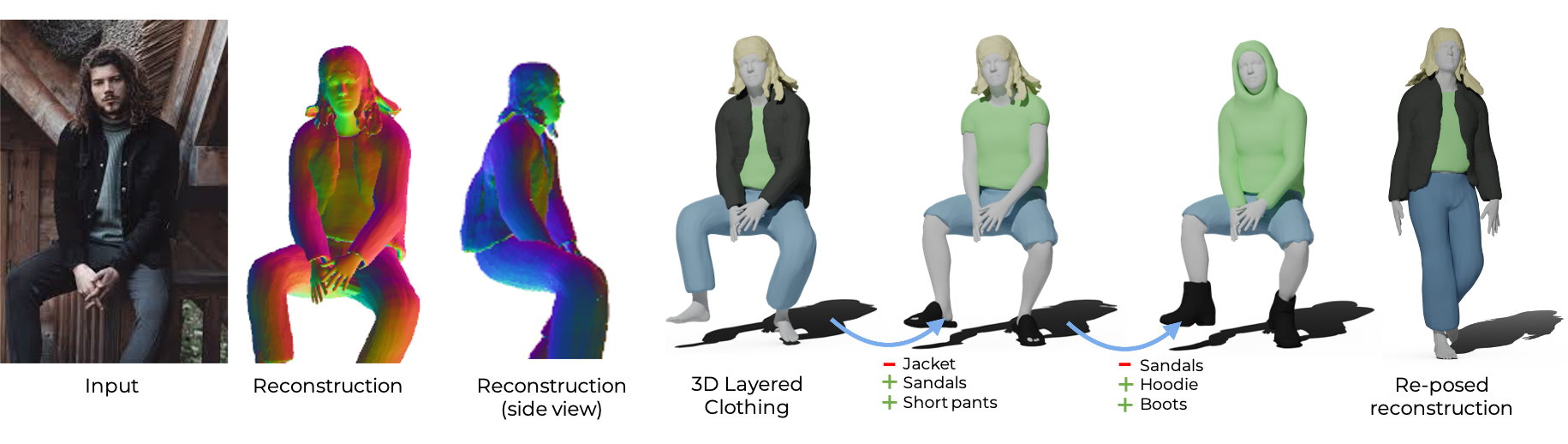}
    \vspace{-2.5mm}
    \captionof{figure}{\small{We introduce SMPLicit, a fully differentiable generative model for clothed bodies, capable of representing garments  with different topology. The four figures on the left show the application of the model to the problem of 3D body and cloth reconstruction from an input image. We are able to predict different models per cloth, even for multi-layer cases. Three right-most images: The model can also be used for editing the outfits, removing/adding new garments and re-posing the body. }}\label{fig:teaser}
\end{center}
}]


\vspace{-1mm}
\begin{abstract}
\vspace{-3mm}

In this paper we introduce SMPLicit, a novel  generative model to  jointly represent  body pose, shape and  clothing geometry. In contrast to existing learning-based approaches that require training specific models for each type of garment, SMPLicit can represent in a unified manner different garment topologies (\eg from sleeveless tops  to hoodies and to open jackets),  while controlling other properties like the garment size or tightness/looseness. We show our model to be applicable to a large variety of garments including T-shirts, hoodies, jackets, shorts, pants, skirts, shoes and even hair.
The representation flexibility of SMPLicit builds upon an implicit model  conditioned with the  SMPL human body parameters and a learnable  latent space which is semantically interpretable and aligned with the clothing attributes.  
The proposed model is fully differentiable, allowing for its use  into  larger end-to-end trainable systems. In the experimental section, we demonstrate \Modelname~can be readily used  for  fitting 3D  scans and for 3D reconstruction in images of dressed people. In both cases we are able to go beyond state of the art, by retrieving complex garment geometries, handling situations with multiple clothing layers and providing a tool for easy outfit editing. To stimulate further research in this direction, we will make our code and model publicly available at \url{http://www.iri.upc.edu/people/ecorona/smplicit/}.


\end{abstract}

\section{Introduction}
Building a differentiable and low dimensional generative model  capable to control   garments style and deformations under different body shapes and poses would open the door to many exciting applications in \eg digital animation of clothed humans, 3D content creation and virtual try-on. However, while such representations have been shown effective for the case of the undressed human body~\cite{smpl,smplx}, where body shape variation can be encoded by a few parameters of a linear model,  there exist so far, no similar approach for doing so on clothes. 

The standard practice to represent the geometry of dressed people has been to treat clothing as an additive displacement over canonical body shapes, typically obtained with SMPL~\cite{tex2shape, bcnet, cape, tailornet}. Nevertheless, these types of approaches cannot tackle the main challenge in garment modeling, which is  the large variability of types, styles, cut, and deformations they can have. For instance, upper body clothing can be either a sleeveless top, a long-sleeve hoodie  or an open jacket. In order to handle such variability, existing approaches need to train 
specific models for each type of garment, hampering thus their  practical utilization.

In this paper, we introduce \Modelname, a topologically-aware generative model for clothed bodies that can be controlled by a low-dimensional and interpretable vector of parameters. 
\Modelname~builds upon an implicit network architecture conditioned on the body pose and shape. With these two factors, we can predict clothing deformation in 3D as a function of the body geometry, while controlling the garment style (cloth category) and cut (\eg sleeve length, tight or loose-fitting). 
We independently train this model for two distinct cloth clusters, namely {\em upper body} (including sleeveless tops, T-shirts, hoodies and jackets) and {\em lower body} (including pants, shorts and skirts).  Within each cluster, the same model is able to represent garments with very different geometric properties and topology while allowing to smoothly and consistently interpolate between their geometries. {\em Shoes} and {\em hair} categories are also   modeled as independent categories.  Interestingly, \Modelname~is fully differentiable and can be easily deployed and integrated into larger end-to-end deep learning systems. 

Concretely, we demonstrate that SMPLicit can be readily applied to two different problems. First,  for fitting 3D scans of dressed people. In this problem, our multi-garment ``generic'' model is on a par with other approaches that were specifically trained for each garment~\cite{cape, tailornet}. We also apply \Modelname~for the challenging problem of 3D reconstruction from images, where we compare favorably to state-of-the-art, being able to retrieve complex garment geometries under different body poses, and can tackle situations with multiple clothing layers. Fig.~\ref{fig:teaser} shows one such example, where besides reconstructing the geometry of the full outfit, \Modelname~provides semantic knowledge of the shape, allowing then  for garment editing and body re-posing, key ingredients of virtual try-on systems.  

To summarize, the main contributions of our work are:  (1) A generative model that is capable of  representing  clothes under different topology; (2) A low-dimensional and semantically interpretable latent vector for controlling clothing style and cut; (3) A model that can be conditioned on human pose, shape and garment style/cut; (4) A fully differentiable model for easy integration with deep learning; (5) A versatile approach that can be applied to both 3D scan fitting and 3D shape reconstruction from images in the wild; (6) A 3D reconstruction algorithm that produces controllable and editable surfaces.

\begin{figure*}[t!]
  \vspace{-0.45cm}
  \includegraphics[width=\linewidth, trim={0.1cm 0.1cm 0cm 0}, clip = true]{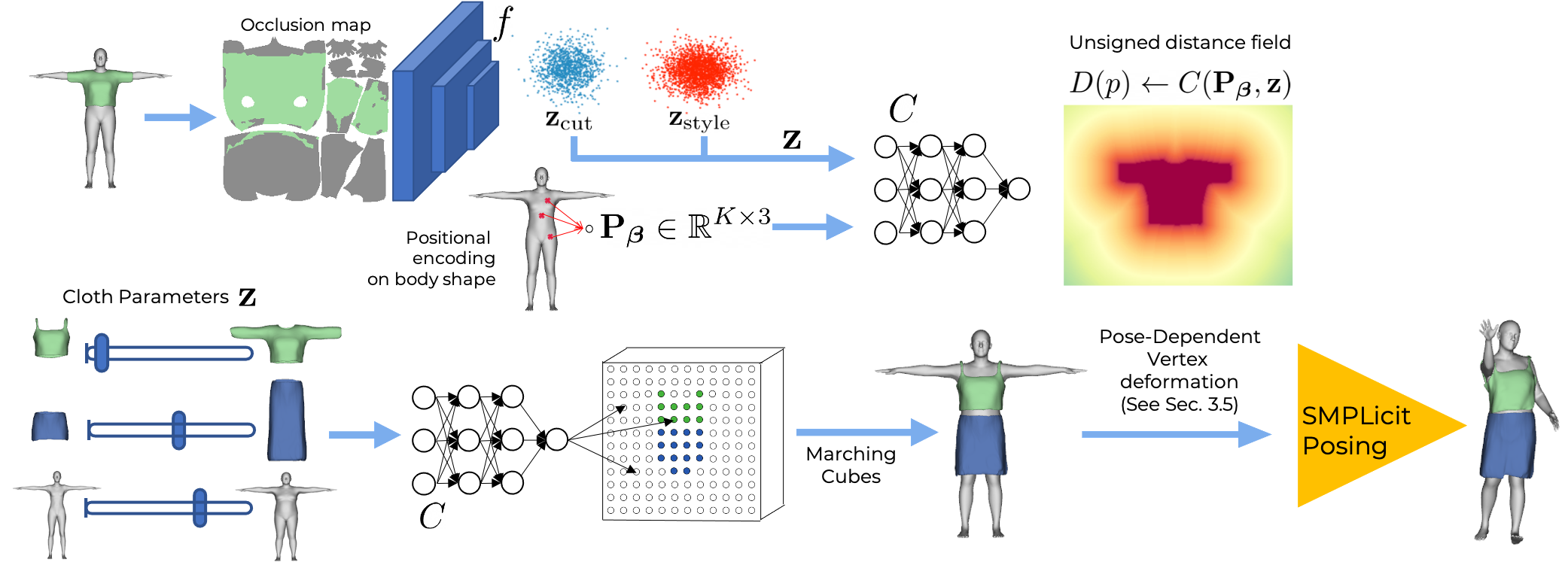}
  \put(-50,160){\underline{\large	{Training}}}
  \put(-55,90){\underline{\large{	Inference}}}
  \vspace{-0.1cm}
  \caption{\small{\textbf{Architecture of \Modelname~during training (top row) and  inference (bottom row)}. 
  At the core of \Modelname~lies an implicit-function network $C$ that predicts unsigned distance from the query point $\mathbf{p}$ to the cloth iso-surface. The input $\posenc$ is encoded from $\mathbf{p}$ given a body shape. During training, we jointly train the network $C$ as the latent space representation is created. We include an image encoder $f$ that takes SMPL occlusion maps from ground truth garments and maps them to shape representations $\cut$, and a second component $\style$ trained as an auto-decoder~\cite{deepsdf}.
  At inference, we run the network $C(\cdot)$ for a densely sampled 3D space and use Marching Cubes to generate the 3D garment mesh. We finally pose each cloth vertex  using the learnt skinning parameters~\cite{smpl} of the closest SMPL vertex.
  }}
  \label{fig:architecture}
  \vspace{-0.4cm}
\end{figure*}

\section{Related work}
Cloth modeling is a long-standing goal lying at the intersection of computer vision and computer graphics. We next discuss related works, grouping them in {\em Generative cloth models} and {\em 3D reconstruction of clothed humans}, the two main topics  in which we contribute.


\subsection{Generative cloth models}



Drawing inspiration on the success of the data driven methods for modeling the human body~\cite{anguelov2005scape,ponsmollSIGGRAPH15Dyna, corona2020ganhand, totalcapture,smpl,smplx,mano}, a number of approaches aim to learn clothing models from real data,  obtained using multiple images~\cite{alldieck2019learning,alldieck2018video,tex2shape, bhatnagar2019multi,habermann2019livecap}, 3D scans~\cite{neophytou2014layered,ponsmollSIGGRAPH17clothcap,bhatnagar2020ipnet}  or RGBD sensors~\cite{doublefusion, yu2019simulcap}. Nevertheless, capturing a sufficiently large volume of data to  represent the complexity of clothes is still an open challenge, and methods built using real data~\cite{corona2018active, yang2018analyzing, deepwrinkles} have problems to generalize beyond the deformation patterns of the training data. ~\cite{cape} addresses this limitation by means of a probabilistic formulation  that predicts clothing displacements on the graph defined by the SMPL mesh. While this strategy improves the generalization capabilities, the clothes it is able to generate can not largely depart from the shape of a  ``naked'' body  defined by SMPL.

An alternative to the use of real   data is to learn clothing models using data    from physics simulation engines~\cite{guan2012drape,gundogdu2019garnet,tailornet,santesteban2019learning,wang2018learning}. The accuracy of these models, however, is again constrained by the quality of the simulations. Additionally, their underlying methodologies still rely on displacement maps from a template, and can not produce different topologies. 

Very recently,~\cite{bcnet,shen2020garmentgeneration,vidaurre2020fully} have proposed  strategies to model garments with topologies departing from the SMPL  body mesh, like skirts or dresses. \cite{bcnet} does so by predicting generic skinning weights for the garment, independent from those of the body mesh. In~\cite{vidaurre2020fully}, the garment is characterized by means of 2D sewing patterns, with a set of parameters that control its 3D shape.  A limiting factor of these approaches is that they   require training specific models for each type of garment, penalizing thus their practical use. ~\cite{shen2020garmentgeneration} uses also sewing patterns to build a unified representation encoding different clothes. This representation, however, is too complex to allow controlling the generation process with just a few parameters. \Modelname, in contrast, is able to represent using a single low-dimensional parametric model a large variety of clothes, which largely differ in  their geometric properties, topology  and cut. 

Table~\ref{tab:related_work} summarizes the main properties of the most recent generative cloth models we have discussed. 



\begin{table}
\centering
\vspace{-4mm}
\resizebox{0.46 \textwidth}{!}{%
\begin{tabular}{c c c c c c}
\cmidrule(lr){1-1} \cmidrule(lr){2-6}
Method &  \multicolumn{1}{c}{\begin{tabular}[c]{@{}c@{}}Body Pose\\ Variations\end{tabular}} & \multicolumn{1}{c}{\begin{tabular}[c]{@{}c@{}}Body Shape\\ Variations\end{tabular}} & Topology & \multicolumn{1}{c}{\begin{tabular}[c]{@{}c@{}}Low-Dimension\\ Latent Vector\end{tabular}} & \multicolumn{1}{c}{\begin{tabular}[c]{@{}c@{}}Model is\\ public\end{tabular}} \\

\cmidrule(lr){1-1} \cmidrule(lr){2-6} 


Santesteban~\cite{santesteban2019learning} & \checkmark & \checkmark & & & \\
DRAPE~\cite{guan2012drape} & \checkmark & \checkmark & & \checkmark & \\
Wang~\cite{wang2018learning}  & & \checkmark & & \checkmark & \checkmark \\
GarNet~\cite{gundogdu2019garnet} & \checkmark & \checkmark & & & \checkmark  \\

TailorNet~\cite{tailornet}& \checkmark & \checkmark& & \checkmark & \checkmark \\
BCNet~\cite{bcnet} & \checkmark & \checkmark & &  \checkmark & \checkmark \\

Vidaurre~\cite{vidaurre2020fully} & \checkmark & \checkmark & &  \checkmark &  \\
Shen~\cite{shen2020garmentgeneration} & \checkmark & \checkmark & \checkmark &  & \checkmark \\

\Modelname &\checkmark & \checkmark & \checkmark & \checkmark & \checkmark
\\
\cmidrule(lr){1-1} \cmidrule(lr){2-6}
\end{tabular}}
\vspace{-0.3cm}
\caption{\small{\bf Comparison of our method with other works.} }
\label{tab:related_work}
\vspace{-0.5cm}
\end{table}

\subsection{Reconstructing  clothed humans from  images}

Most approaches for reconstructing 3D humans from images return the SMPL  parameters, and thus only retrieve 3D body meshes, but not clothing~\cite{bogo2016keep,guan2009estimating,hmr,kolotouros2019spin,kolotouros2019convolutional, lassner2017unite,omran2018neural,smplx,frankmocap,smith2019facsimile,xu2020humanshape}. To reconstruct clothed people, a standard practice is to represent clothing geometry as an offset over the SMPL body mesh~\cite{alldieck2019learning, alldieck2018detailed, alldieck2018video,onizuka2020tetratsdf,tex2shape,bhatnagar2019multi,lazova3dv2019,sayo2019human,zhu2019detailed}. However, these  approaches are prone to fail for loose garments that exhibit large displacements over the body. 

Non-parametric representations have also been explored for reconstructing arbitrary clothing topologies. These include approaches based on volumetric voxelizations~\cite{varol2018bodynet}, geometry images~\cite{pumarola20193dpeople}, bi-planar depth maps~\cite{gabeur2019moulding} or  visual hulls~\cite{natsume2019siclope}. Certainly, the most powerful    model-free representations   are those based on implicit functions~\cite{saito2019pifu,saito2020pifuhd,chibane20ifnet}. Recent approaches have also combined parametric and model-free representations, like SMPL plus voxels~\cite{zheng2019deephuman} and SMPL plus implicit functions~\cite{bhatnagar2020ipnet,huang2020arch}.

While these approaches retrieve rich geometric detail, the resulting surfaces can not be controlled in both pose and clothing. \Modelname~is also built upon implicit functions, but our output contains multiple layers for the body and garments, and allows control over pose and clothing.

\section{\Modelname}\label{sec:approach}

We next describe the \Modelname~formulation, training scheme and how it can be used to interpolate between clothes. Fig.~\ref{fig:architecture} shows the whole train and inference process.


\subsection{Vertex Based SMPL vs SMPLicit}

We build on the parametric human model SMPL~\cite{smpl} to generate clothes that adjust to a particular human body $M(\shape, \pose)$, given its shape $\shape$ and pose $\pose$. 
SMPL is a function 
\begin{equation}
    M (\shape, \pose): \pose \times{\shape} \mapsto \mat{V} \in \mathbb{R}^{3N} \label{eq:posing},
\end{equation}
which predicts the $N$ vertices $\mat{V}$ of the body mesh as a function of pose and shape. Our goal is to add a layer of clothing on top of SMPL. Prior work adds displacements~\cite{alldieck2019learning,alldieck2018video} on top of the body, or learns garment category-specific vertex-based models~\cite{gundogdu2019garnet,tailornet}. The problem with predicting a fixed number of vertices is that different topologies (T-shirt vs open jacket) and extreme geometry changes (sleeve-less vs long-sleeve) can not be represented in a single model.

Our main contribution is \emph{SMPLicit-core} (Sec.\ref{subsec:smplicit_core}-\ref{subsec:inference}), which departs from vertex models, and predicts clothing on  T-pose with a learned implicit function
\begin{equation}
C(\mat{p}, \shape, \cut, \style) \mapsto \mathbb{R}^{+}.
\label{eq:smplicit_core}
\end{equation}
Specifically, we predict the \emph{unsigned distance} to the clothing surface for a given point $\mat{p} \in \mathbb{R}^3$.
By sampling enough points, we can reconstruct the desired mesh by thresholding the distance field and running Marching Cubes~\cite{marching_cubes}.
In addition to shape, we want to control the model with intuitive parameters $(\cut,\style)$ representing the \emph{cut} (\eg, long vs short) and \emph{style} (\eg, hoodie vs not hoodie) of the clothing. 
Moreover, although it is not the focus of this paper, we also learn a point-based displacement field (Sec.\ref{subsec:pose_dependent}) to model pose-dependent deformations, and use SMPL skinning to pose the garments. The full model is called \Modelname~ and outputs posed meshes $\mathcal{G}$ on top of the body:
\begin{equation}
C^\prime(\pose, \shape, \cut, \style) \mapsto \mathcal{G}.
\label{eq:smplicit}
\end{equation}





\subsection{\Modelname-Core Formulation}
\label{subsec:smplicit_core}

We explain here how we  learn the \emph{input representation}: two latent spaces to control clothing cut and style, and body shape to control fit; and the \emph{output representation}. Together, these representations allow to generate and control garments of varied topology in a single model. 

\vspace{1mm}
\noindent{\bf Clothing cut:} We aim  to control the output clothing cut, which we define as the body area  occluded by clothing. To learn a latent space of cut, for each garment-body pair in the training set, we compute a UV body \emph{occlusion image} denoted as $\mat{U}$. That is, we set every pixel in the SMPL body UV map to $1$ if the corresponding body vertex is occluded by the garment, and $0$ otherwise, see Fig.~\ref{fig:architecture}. Then we train an image encoder $f:\mat{U} \mapsto \cut \in \mathbb{R}^D$ to map the occlusion image to a latent vector $\cut$. 

\vspace{1mm}
\noindent{\bf Clothing style:} Different clothes might have the same body occlusion image $\mat{U}$, but their geometry can differ in tightness, low-frequency wrinkles or collar details. 
Thus we add another subset of parameters $\mat{z}_c$ which are initialized as a zero-vector and trained following the auto-decoder procedure from~\cite{deepsdf}.

The set of parameters $\mat{z} = [\cut, \style] \in \mathbb{R}^{N}$ fully describes a garment cut and style.

\vspace{1mm}
\noindent{\bf Body shape:} Since we want the model to vary with body shape, instead of learning a mapping from points to occupancy~\cite{chen2019learning,mescheder2019occupancy, deepsdf}, we first encode points relative to the body.
For each garment, we identify SMPL vertices that are close to ground truth models (\eg torso vertices for upper-body clothes), and obtain $K$ vertex clusters $\mat{v}_k \in \mathbb{R}^3$ that are distributed uniformly on the body in a T-pose.
Then we map a 3D point in space $\mat{p} \in \mathbb{R}^3$ to a body relative encoding $\mat{P}_\shape \in \mathbb{R}^{K\times 3}$ matrix, with rows storing the displacements to the clusters $\mat{P}_{\shape,k} = (\mat{p}-\mat{v}_k)$.
This over-parameterized representation allows the network to reason about body boundaries, and we empirically observed superior performance compared to Euclidean or Barycentric distances.

\vspace{1mm}
\noindent{\bf Output representation:}
One of the main challenges in learning a 3D generative clothing model is registering training garments~\cite{bhatnagar2019multi,ponsmollSIGGRAPH17clothcap} (known to be a hard problem), which is necessary for vertex-based models~\cite{cape,tailornet}. Implicit surface representations do not require registration, but necessitate closed surfaces for learning occupancies~\cite{mescheder2019occupancy, pumarola2020d} or signed distances~\cite{deepsdf, li2020frodo}.
Since garments are open surfaces, we follow recent work~\cite{chibane2020neural} by predicting unsigned distance fields.

Given a query point $\mat{p}$, its positional encoding $\posenc$ and cloth parameters $\mat{z}$, we train a decoder network $C(\posenc, \mat{z})\mapsto\mathbb{R}^+$ to predict the unsigned distance $D(\mat{p})$ to the ground truth cloth surface.

\subsection{\Modelname-core Training}
Training entails learning the network parameters $\mat{w}_1$ of the clothing cut image encoder $\cut = f(\mat{U};\mat{w_1})$, the style latent parameters $\style$ for each training example, and the parameters of the decoder network $C(\cdot;\mat{w}_2)$.
For one training example, and one sampled point $\mat{p}$, we have the following loss:
\begin{equation}
    \mathcal{L}_d = |C(\posenc, f(\mat{U};\mat{w}_1),\style; \mat{w}_2) - D(\mat{p}) |.
\end{equation}
During training, we sample points uniformly on a body bounding box, and also near the ground-truth surface, and learn a model of \emph{all} garment categories jointly (we train separate models for upper-body, pants, skirts, shoes and hair though, because interpolation among them is not meaningful).
At inference, we discard the encoder $f:\mat{U}\mapsto \cut$ network, and control SMPLicit directly with $\cut$. 

To smoothly interpolate and generate new clothing, we constrain the latent space $\mat{z} = [\cut,\style]$ to be distributed normally with a second loss component $\mathcal{L}_z = | \mat{z}|$. 

We also add zero mean identity covariance Gaussian noise $\mat{z}_\sigma \sim \mathcal{N}(\mat{0},\sigma_n\mat{I})$ in the cloth representations before the forward pass during training, taking as input $C(\posenc, \mat{z} + \mat{z}_\sigma)$, which proves specially helpful for garment types where we have a very small amount of data.
The network $C$ and the cloth latent spaces are jointly learned by minimizing a linear combination of the previously defined losses $\mathcal{L}_d + \lambda_z \mathcal{L}_z$, where $\lambda_z$ is a  hyper-parameter.

\subsection{\Modelname-core Inference}
\label{subsec:inference}
To generate a 3D garment mesh, we evaluate our network $C(\cdot)$ at densely sampled points around the body in a T-pose, and extract the iso-surface of the distance field at threshold $t_d$ using Marching Cubes~\cite{marching_cubes}. We set the hyperparameter $t_d=0.1\,mm$ such that reconstructed garments do not have artifacts and are smooth.
Since $C(\cdot)$ predicts unsigned distance and $t_d > 0$, the reconstructed meshes have a slightly larger volume than ground truth data; this is still better than closing the garments for training which requires voxelization. 
Thinner surfaces could be obtained with Neural Distance Fields~\cite{chibane2020neural}, but we leave this for future work.

In summary, we can generate clothes that fit a body shape $\shape$ by: (1) sampling $\mat{z} \sim\ \mathcal{N}(\mu*\mat{1}, \sigma*\mat{I})$, with a single mean and variance $(\mu,\sigma \in \mathbb{R})$ for all latent components obtained from the training latent spaces; (2) estimating the positional encoding $\posenc$ for points around the T-pose and evaluating 
$C(\posenc ,\mat{z})$; (3) thresholding the distance field, and (4) running marching cubes to get a mesh. 


\begin{figure}[t!]
  \vspace{-0.5cm}
  \includegraphics[width=1.\linewidth, trim={0cm 0cm 2.8cm 0}, clip = true]{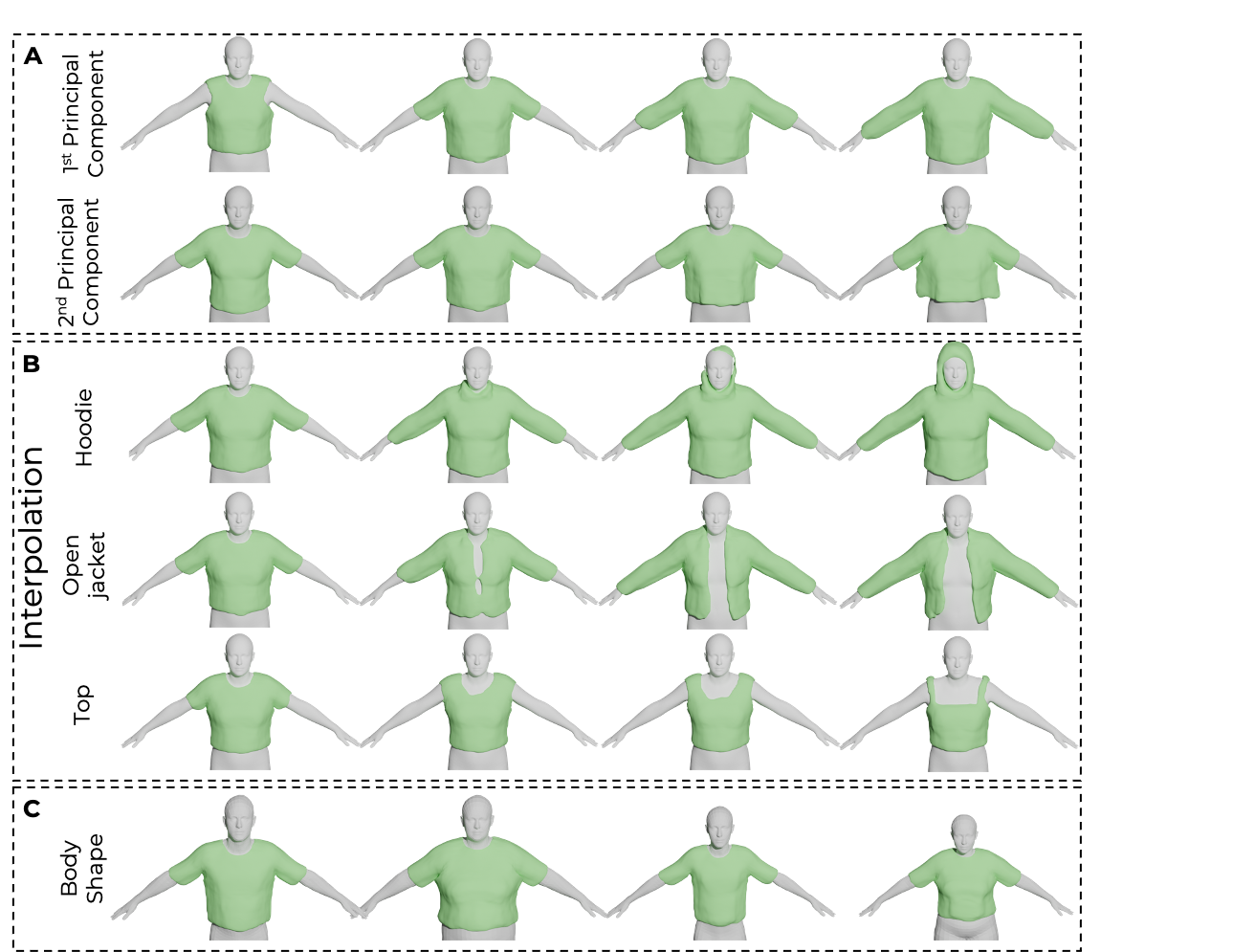}
  \vspace{-0.5cm}
  \caption{\small{\textbf{Overview of interpolations on latent space.}
  (A) effect of the two first principal components in the garment geometry. 
  (B) \Modelname~can be used to interpolate from T-shirts to more complex clothes like hoodies, jackets or tops. 
  (C) examples of retargeting an upper-body cloth to different human body shapes. }
  }
  \label{fig:qualitative_generation}
  \vspace{-0.4cm}
\end{figure}
\subsection{Pose Dependent Deformation}
\label{subsec:pose_dependent}

\Modelname-core can drape garments on a T-posed SMPL, but does not predict pose dependent deformations. Although \emph{pose deformation is not the focus} of this work, we train a pose-dependent model to make SMPLicit readily available for animation applications. Similar to prior work~\cite{tailornet}, we learn the pose-deformation model on a canonical T-pose, and use SMPL learned skinning to pose the deformed mesh.
Here, we leverage the publicly available TailorNet~\cite{tailornet} dataset of simulated garments. Specifically, we learn a second network which takes body pose $\pose$, a learnable latent variable $\zpose$
and maps them to a per-point displacement $P: \mat{p} \times \pose \times \zpose \mapsto \mat{d} \in \mathbb{R}^3$. The latent space of $\zpose$ is learned in an auto-decoding fashion like $\style$. 

During training, since we are only interested in the displacement field on the surface, we only evaluate the model on points sampled along the cloth surface template on a T-Pose. 
We also encode the position of the input points $\mat{p}\mapsto\mat{P}_\shape$ as a function of the body surface and train the model to minimize the difference between ground truth displacement and prediction.

During inference, we only evaluate $P$ on the vertices of the recovered \Modelname-core mesh, and displace them accordingly $\mat{p}\mapsto\mat{p}+\mat{d}$ to obtain a deformed mesh (still in the T-pose).
Then we apply SMPL~\cite{smpl} to both body and deformed garment to pose them with $\pose$.
In particular, we deform each garment vertex using the skinning deformation of the closest SMPL body vertex. This process determines the \Modelname~ function $C^\prime(\cdot)$ defined in Eq.~\eqref{eq:smplicit}.

\begin{figure*}[t!]
 \vspace{-0.6cm}
  \includegraphics[width=.99\linewidth, trim={0 0 0.1cm 0}, clip = true]{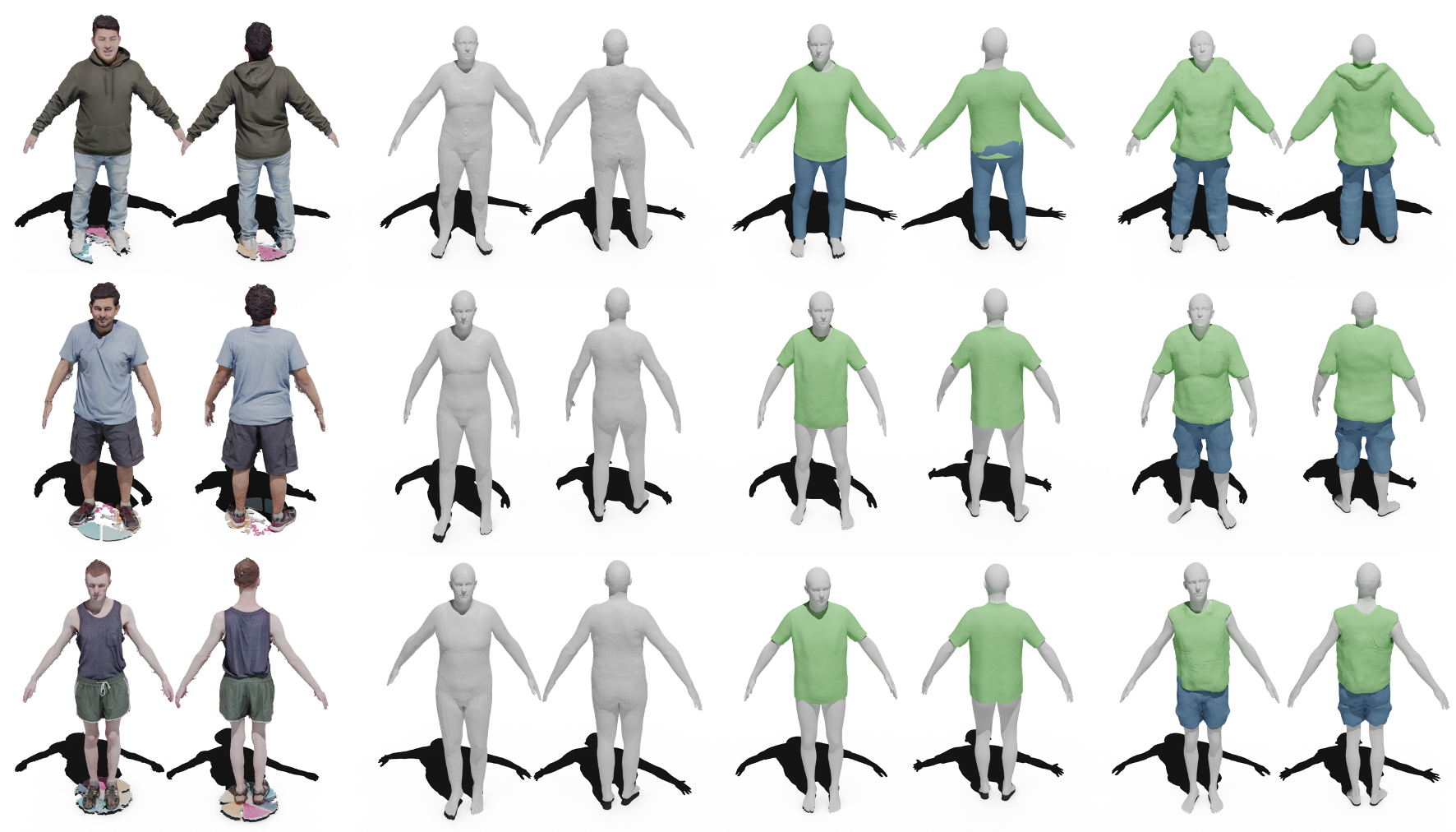}\\
  \put(43, 5){3D Scans}
  \put(167, 5){CAPE~\cite{cape}}
  \put(287, 5){TailorNet~\cite{tailornet}}
  \put(417,5){\Modelname}
  \vspace{-0.15cm}
  \caption{\small{\textbf{Fitting \Modelname~to 3D Scans of the Sizer Dataset~\cite{sizer}}. All three models achieve fitting results of approximately 1 mm of error.  However, \Modelname~does this using a single model that can represent varying clothing topologies. For instance, it can model either hoodies (top row) and  tank tops (third row) or long and short pants.}}
  \label{fig:fitting_scans}
  \vspace{-0.4cm}
\end{figure*}
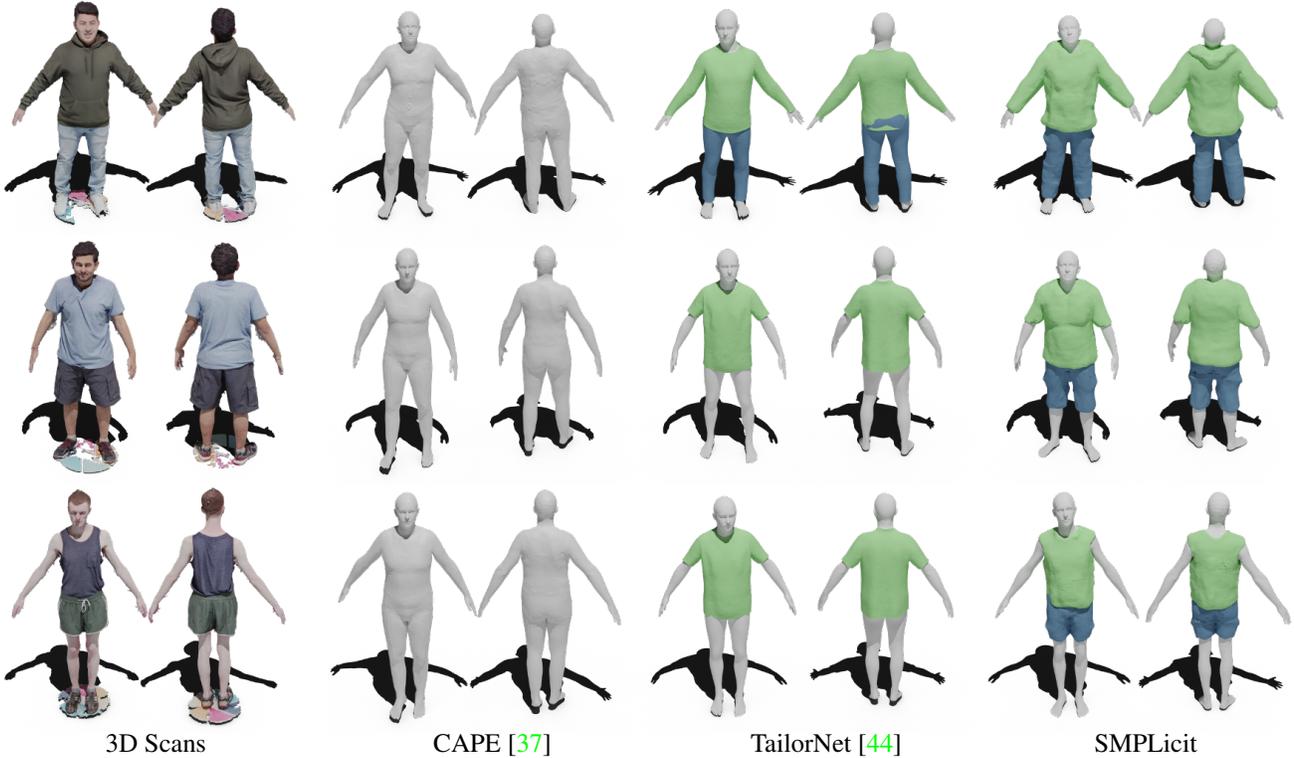

\begin{figure*}[t!]
  \vspace{-0.5cm}
  \includegraphics[width=.99\linewidth, trim={0.05cm 0.5cm 0.3cm 0.01cm}, clip = true]{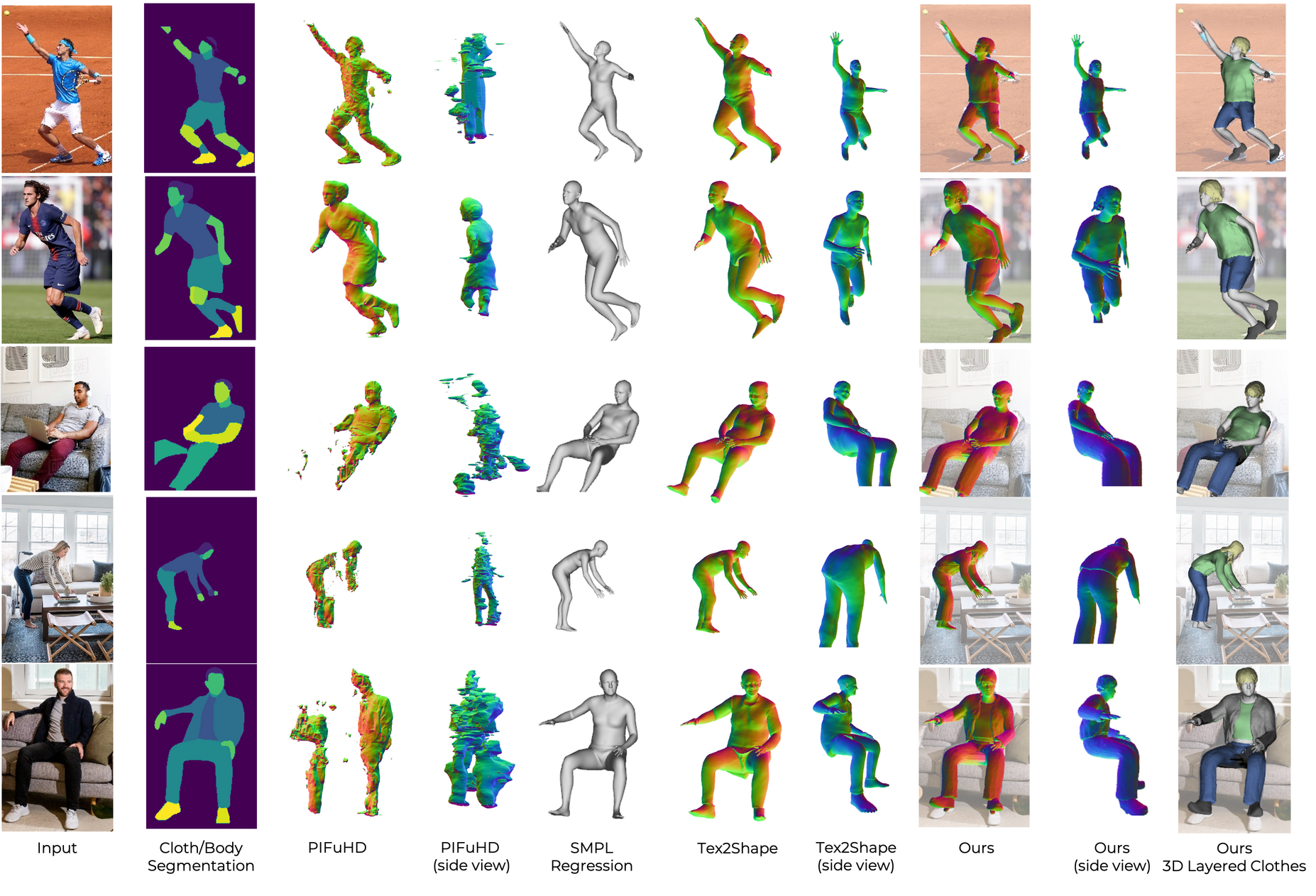}
  \vspace{-0.3cm}
   \caption{\small{\textbf{3D reconstruction of clothed humans,} 
   in comparison to PIFuHD~\cite{saito2020pifuhd} and Tex2Shape~\cite{tex2shape}.
   SMPL regression is  from~\cite{frankmocap}.}}
  \label{fig:fitting_images}
  \vspace{-0.4cm}
\end{figure*}

\section{Applications of \Modelname}

In this section, we show the potential of \Modelname~ for several computer vision and graphics applications. We demonstrate how to interpolate garments in the latent space and edit their cut and style.
We then show how \Modelname~can be fitted to 3D scans of dressed humans, or directly to in-the-wild images for perception tasks, taking advantage of the full differentiability of the predicted unsigned distance field with respect to cloth parameters.

\subsection{Generative properties}
To provide control to the user, we perform PCA on the latent space to discover directions which vary intuitive cloth properties, like sleeve-length, and identify cloth prototypes such as hoodies and tops.

\vspace{1mm}
\noindent{\bf PCA:} The latent space  $\mat{z} = [\cut,\style]$  of \Modelname-core is small ($4$ to $18$) in order to better disentangle cloth characteristics. 
We further perform PCA on the $\cut$ latent space and find that, for the upper and lower-body clothes, the first component controls sleeve length, while the second changes overall length (for upper-body garments),
or the waist boundary height (for pants and skirts). Fig.~\ref{fig:qualitative_generation}-(A) shows the effect of the first 2 components for upper-garment. 
We also notice that perfect disentanglement from cut and style is not possible, as for example the network learns that tops tend to be more loose than t-shirts.  

\vspace{1mm}
\noindent{\bf Prototypes:} Furthermore, we identify cloth prototypes with interesting characteristics in the train data, such as open jackets, hoodies or tops, and store their average style latent space vectors $\mat{z}$. 
Fig.~\ref{fig:qualitative_generation}-(B) illustrates interpolation from a T-shirt to each of these prototypes; notice how SMPLicit is able to smoothly transition from short-sleeve to open jacket.

\vspace{1mm}
\noindent{\bf Body Shape:} In Fig.~\ref{fig:qualitative_generation}-(C), we show results of re-targeting a single T-Shirt to significantly different body shapes. 

\begin{table}
\vspace{-4mm}
\centering
\resizebox{0.46 \textwidth}{!}{%
\begin{tabular}{c c c c c}
\cmidrule(lr){1-1} \cmidrule(lr){2-3} \cmidrule(lr){4-5}
& \multicolumn{4}{c}{Distance to surface (mm)} \\
& \multicolumn{2}{c}{Short Sleeves} & \multicolumn{2}{c}{Long Sleeves}  \\
Method & Lower-Body & Upper-Body & Lower-Body & Upper-Body \\

\cmidrule(lr){1-1} \cmidrule(lr){2-3} \cmidrule(lr){4-5}
Cape~\cite{cape} & 1.15 & 0.87 & 1.09 & 1.35 \\
TailorNet~\cite{tailornet} & - & 0.32 & 0.48 & 0.41 \\
\Modelname & 0.78 & 0.46 & 0.58 & 0.52\\

\cmidrule(lr){1-1} \cmidrule(lr){2-3} \cmidrule(lr){4-5}

\end{tabular}}
\vspace{-2mm}
\caption{\small{{\bf Capacity of \Modelname~for fitting 3D scans in comparison with TailorNet~\cite{tailornet} and CAPE~\cite{cape}}.
Note that we fit clothes on either long-sleeves or short-sleeves using a single model, while baselines have particularly trained for such topologies. All models achieve a remarkably accurate fitting within the segmented clothes of the original 3D scans.}}
\label{tab:quantitative_scans}
\vspace{-5mm}
\end{table}
\subsection{Fitting \Modelname~to 3D scans of dressed people}

Here we show how to fit \Modelname~to 3D scans of the Sizer dataset~\cite{sizer} which includes cloth segmentation.
Intuitively, the main objective for fitting is to impose that \Modelname-core evaluates to zero at the \emph{unposed} scan points. 
We sample 3D points uniformly on the segmented scan upper-body and lower-body clothes, and also the 3D empty space around it. 
Let $\mat{q}\in\mathbb{R}^3$ be a point in the posed scan space, and let $\mat{d} = \mathrm{dist}(\mat{q},\mathcal{S})$ be the distance to the scan. 
Since \Modelname-core is defined on the T-pose, we unpose $\mat{q}$ using the differentiable SMPL parameters (we associate to the closest SMPL vertex), 
and obtain the body relative encoding $\posenc (\pose,\shape)$, now as a function of shape \emph{and} pose.
Then we impose that our model $C$ evaluates to the same distance at the encoding of the unposed point:
\begin{equation}
\mathcal{L}(\shape,\pose,\mathbf{z})= |C(\posenc(\pose,\shape), \mathbf{z}) - \mat{d}| .
\end{equation}
We run the optimization for a number of iterations and for the cloth parameters of all garments the person is wearing. We also minimize the Chamfer distance between scan points and SMPL vertices, the MSE between SMPL joints and predicted scan joints, an SMPL prior loss~\cite{bogo2016keep}, and a regularization term for $\boldsymbol{z}$.
We use scheduling and first optimize the pose and shape, and finally all parameters jointly. See the Supp. Mat. for more details.

\subsection{Fitting \Modelname~to images} \label{sec:fitimages}

Similar to SMPL for undressed bodies, \Modelname~provides the robustness and semantic knowledge to reconstruct clothed people in images, especially in presence of severe occlusions, difficult poses, low-resolution images and noise.
We first detect people and obtain an estimate of each person's pose and shape~\cite{frankmocap}, as well as a 2D cloth semantic segmentation~\cite{rp-r-cnn}.
We then fit \Modelname~to every detection to obtain layered 3D clothing. 

For every detected garment, we uniformly sample the space around the T-Posed SMPL, deform those points to the target SMPL pose $(\mat{p}\mapsto \bar{\mat{p}})$, and remove those that are occluded by the own body shape.
Each \emph{posed} point $\bar{\mat{p}}$ is then projected, falling into a semantic segmentation pixel $(u,v)$ that matches its garment class $s_{\mat{p}} = 1$ or another class/background $s_{\mat{p}} = 0$. 
We have the following loss for a single point $\mat{p}$:
\begin{equation}
\mathcal{L}_I(\mat{z}) =
\begin{cases}
|C(\posenc,\mat{z})-\mat{d}_{\mathrm{max}}|, & \mbox{if } s_{\mat{p}} = 0 \\
\min_{i}|C(\posenc^i,\mat{z})|, & \mbox{if } s_{\mat{p}} = 1 \\
\end{cases}
\end{equation}
When $s_{\mat{p}} = 0$ we force our model to predict the maximum cut-off distance $\mat{d}_{\mathrm{max}}$ of our distance fields (we force the point to be off-surface). 
When $s_{\mat{p}} = 1$ we force prediction to be zero distance (point in surface). 
Since many points $\bar{\mat{p}}^i$ (along the camera ray) might project to the same pixel $(u,v)$, we take the $\min_{i}(\cdot)$ to consider only the point with minimum distance (closest point to the current garment surface estimate).
Experimentally, this prevents thickening of clothes, which helps when we reconstruct more than one cloth layer.
We also add a regularization loss $\mathcal{L}_{z}=|\mat{z}|$ and optimize it jointly with $\mathcal{L}_I$.


%





\section{Implementation details}\label{sec:implem}
We next describe the main implementation details. Further information is provided in the Suppl. Material and in the code that will be made publicly available. 

For the cloth latent space, we set $|\mat{z}| = 18$ for upper-body, pants, skirts, hair and $|\mat{z}| = 4$ for shoes; the pose-dependent deformation parameters $|\mat{z}_\theta| = 128$, number of positional encoding clusters $K=500$ and iso-surface threshold $t_d=0.1$ mm. We clip the unsigned distance field $d_{max}=10$mm. The implicit network architecture uses three 2-Layered MLPs that separately encode $\cut$, $\style$ and $\mat{P}_\shape$ into an intermediate representation before a last 5-Layered MLP predicts the target unsigned distance field.
\Modelname~is trained using Adam~\cite{adam}, with  an initial learning rate  $10^{-3}$, $\beta_1=0.9$, $\beta_2=0.999$ for 1M iterations with linear LR decay after 0.5M iterations. We use $BS=12$, $\sigma_n = 10^{-2}$ and refine a pre-trained ResNet-18~\cite{resnet} as image encoder $f$. As~\cite{deepsdf}, we use weight normalization~\cite{salimans2016weight} instead of batch normalization~\cite{ioffe2015batch}.

\section{Experiments}\label{sec:exp}

This section first describes the  datasets used to train \Modelname, and then we show results for  fitting 3D scans and 3D reconstruction of dressed people from images.

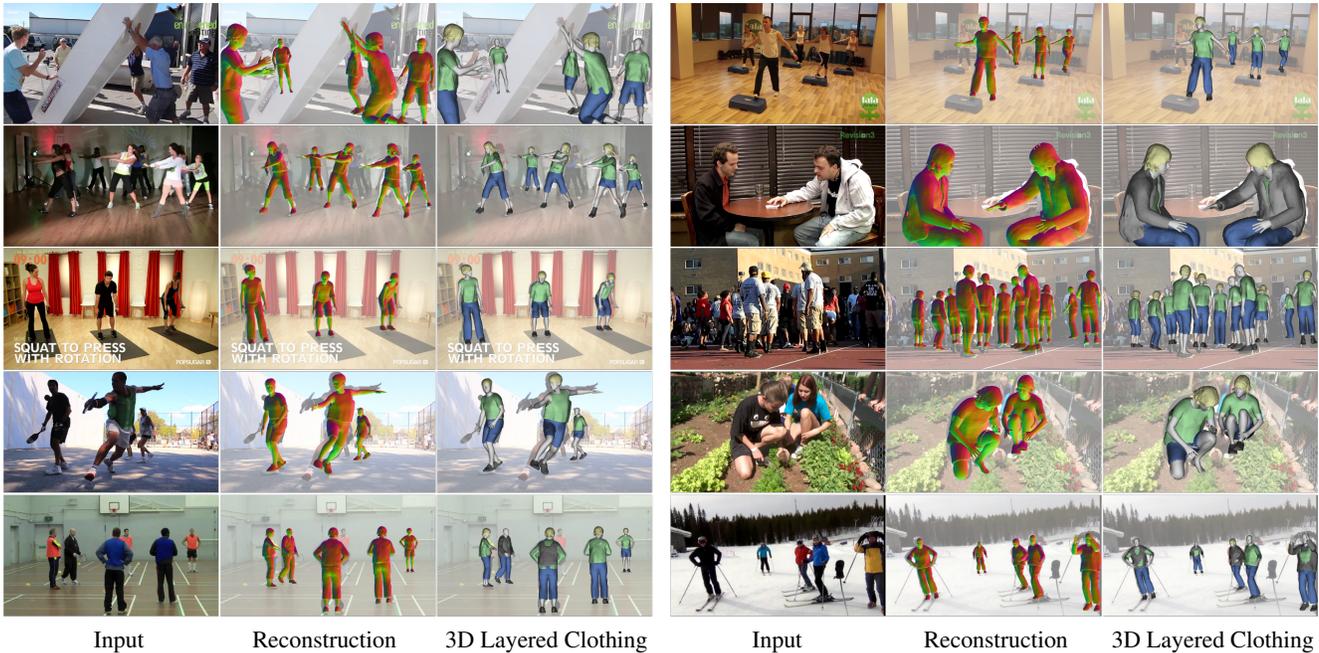
\begin{figure*}[t!]
  \vspace{-0.4cm}
  \includegraphics[width=\linewidth]{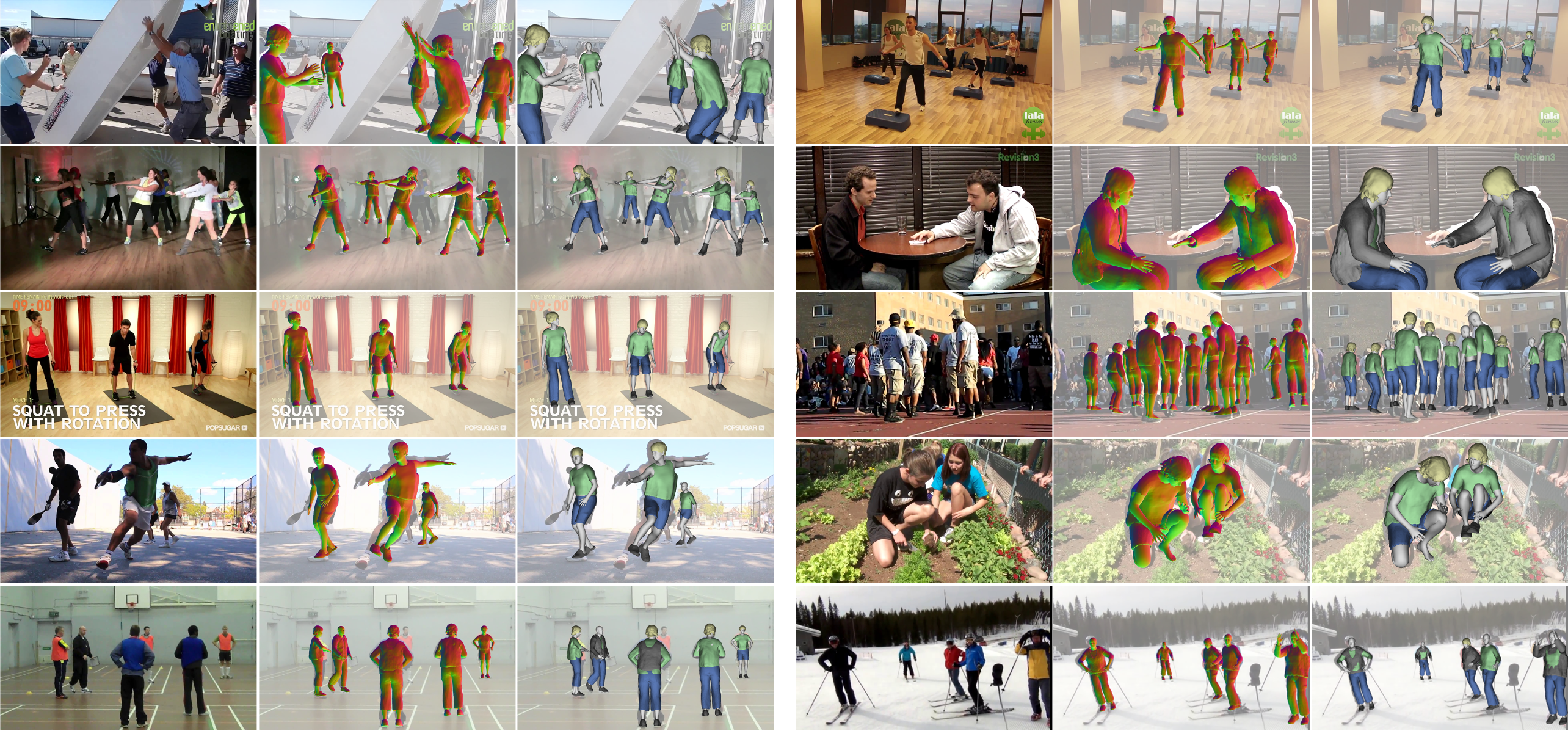}\\
  \put(34,0){\small{Input}}
  \put(94,0){\small{Reconstruction}}
  \put(167,0){\small{3D Layered Clothing}}
  \put(283,0){\small{Input}}
  \put(348,0){\small{Reconstruction}}
  \put(419,0){\small{3D Layered Clothing}}
  \vspace{-0.2cm}
  \caption{\small{\textbf{Fitting \Modelname~in multi-person images from the MPII~\cite{mpii} dataset.} \Modelname~can dress SMPL with a variety of clothes. Failure case in bottom-right example, where cloth semantic segmentation mixes shirts and jackets in most upper-bodies, and \Modelname~ wrongly optimizes two similar intersecting jackets. Best viewed in color with zoom.}}
  \label{fig:fitting_images_multipeople}
  \vspace{-0.4cm}
\end{figure*}

\subsection{Training data}
In order to train \Modelname~we resort to several publicly available datasets and augmentations. Concretely, we use the long-sleeved T-shirts (88797), pants (44265) and skirts (44435) from the BCNet Dataset~\cite{bcnet}. This data is augmented by manually cutting different sleeve sizes on Blender~\cite{blender}, yielding a total of 800k T-shirts, 973k pants and 933k skirts. We also use 3D cloth models of jackets (23), jumpers (6), suits (2), hoodies (5), tops (12), shoes (28), boots (3) and sandals (3) downloaded from diverse public links of the Internet. We adjust these garments to a canonical body shape $\beta = \vec{0}$  and transfer them to randomly sampled body shapes during training, deforming each  vertex using the shape-dependent displacement of the closest SMPL body vertex. For hair, we use the USC-HairSalon dataset~\cite{hairdataset}, which contains 343 highly dense hair pointclouds, mostly of long hair. Given the large imbalance on the cloth categories for the upper-body, in each train iteration we sample one of the downloaded models with probability $0.5$, otherwise we used one of the BCNet garments.

For training the pose-dependent deformation model of Sec.~\ref{subsec:pose_dependent}, we use cloth 
simulations from TailorNet~\cite{tailornet}, which consist of 200  shirt and pants instances. For the remaining garments, except for shoes (which we do not further deform), we train a deformation model parameterized only by $\mat{z}_\theta$, given manually warps generated using  Blender.

\subsection{Fitting \Modelname~to scans of dressed people}
We applied \Modelname-core to the problem of fitting 3D scans of clothed humans from the Sizer dataset~\cite{sizer}, comparing against the recent TailorNet~\cite{tailornet} and  CAPE~\cite{cape}. Since these   methods have been specifically trained for  long-sleeved and short-sleeved (for both shirt and pants), we only evaluate the performance of \Modelname~on these garments.

In Table~\ref{tab:quantitative_scans} we report the reconstruction error (in mm) of the three methods. Note that in our case, we use a single model for modeling both short- and long-sleeves garments, while the other two approaches train independent models for each case. In any event, we achieve results which are comparable to Tailornet, and significantly better than CAPE. Qualitative results of this experiment are shown in Fig.~\ref{fig:fitting_scans}. Note that CAPE does not provide specific meshes for the clothes, and only deforms SMPL mesh vertices. Tailornet yields specific meshes for shirts and long pants. \Modelname, on the other hand,  allows representing different topologies with a single model, from hoodies (first row) to a tank top (third row).

\subsection{3D reconstruction of clothed humans}
Finally, using the optimization pipeline detailed in Sec.~\ref{sec:fitimages}, we  demonstrate that \Modelname~can also be fitted to images of clothed people and provide a 3D reconstruction of the body and clothes. Recall that to apply our method, we initially use~\cite{frankmocap} to estimate SMPL parameters and~\cite{rp-r-cnn} to obtain a pixel-wise segmentation of gross clothing labels (\ie upper-clothes, coat, hair, pants, skirts and shoes).

In Fig.~\ref{fig:fitting_images} we show the results of this fitting on several images in-the-wild with  a single person under arbitrary poses. We compare against PIFuHD~\cite{saito2020pifuhd} and Tex2Shape~\cite{tex2shape}. Before applying PIFuHD,  we  automatically remove the background using~\cite{removebg}, as PIFuHD was trained with no- or simple backgrounds. Tex2Shape  requires  DensePose~\cite{densepose} segmentations, that map input pixels to the SMPL model. As shown in the Figure, the results of \Modelname~consistently improve other approaches, especially PiFuHD, which fails for poses departing from an upright position. Tex2Shape yields remarkably realistic results, but is not able to correctly retrieve the geometry of all the garments. Observe for instance, the example in the last row, where \Modelname~is capable of reconstructing clothing at different layers (T-shirt and jacket). 
Interestingly, once the reconstruction is done, our approach can be used as a virtual try-on,   changing garments' style and reposing the person's position. In Fig.~\ref{fig:teaser} we show one such example.

In Fig.~\ref{fig:fitting_images_multipeople} we go a step further, and show that \Modelname~can also be applied on challenging scenarios with multi-persons, taken from the MPII Dataset~\cite{mpii}. For this purpose we  iterate over all SMPL detections~\cite{frankmocap}, project the body model onto the image and mask out other people's segmentation. Note that in these examples, the model has to tackle extreme occlusions, but the combination of \Modelname~with powerful body pose detectors, like~\cite{frankmocap}, and cloth segmentation algorithms, like~\cite{removebg}, makes this task feasible. Of course, the overall success depends on each individual algorithm. For instance, in the bottom-right example of Fig.~\ref{fig:fitting_images_multipeople}, errors in the segmentation labels are propagated to our reconstruction algorithm which incorrectly predicts two upper-body garments for certain individuals.

\section{Conclusion}
We have presented \Modelname, a generative model for clothing able to  represent different garment topologies and controlling their style and cut with just a few interpretable parameters. Our model is fully differentiable, making it possible to be integrated in several computer vision tasks. For instance,  we  showed that it can be readily used to fit 3D scans, and reconstruct  clothed humans in  images that pose a number of challenges, like multi-layered garments or strong body occlusions due to the presence of multiple people. Additionally, our generative model can be used in geometric content edition tasks to \eg dynamically change the type of garment attributes, opening the door to build novel virtual try-on systems.





\noindent{\bf Acknowledgements}: 
\small{
This work is supported in part by an Amazon Research Award  and by the Spanish government with the projects
HuMoUR TIN2017-90086-R and María de Maeztu Seal of Excellence MDM-2016-0656. Gerard Pons-Moll is funded by the Deutsche Forschungsgemeinschaft (DFG, German Research Foundation) - 409792180 (Emmy Noether
Programme, project: Real Virtual Humans)
}

{\small
\bibliographystyle{ieee_fullname}
\bibliography{references.bib}
}

\end{document}